\documentclass{article}

\usepackage{arxiv}

\usepackage[utf8]{inputenc} 
\usepackage[T1]{fontenc}    
\usepackage{hyperref}       
\usepackage{url}            
\usepackage{booktabs}       
\usepackage{amsmath}
\usepackage{amsfonts}       
\usepackage{bm}
\usepackage{nicefrac}       
\usepackage{microtype}      
\usepackage{lipsum}		
\usepackage{graphicx}
\usepackage{doi}

\title{A modified model for topic detection from a corpus and a new metric evaluating the understandability of topics}

\author{ {\hspace{1mm}Tomoya Kitano} \\
	Graduate School of Information \\ Science and Technology\\
	Osaka University\\
	\texttt{kitano@cas.cmc.osaka-u.ac.jp} \\
	\And
	{\hspace{1mm}Yuto Miyatake} \\
	Cybermedia Center\\
	Osaka University\\
	\texttt{yuto.miyatake.cmc@osaka-u.ac.jp} \\
    \And
    {\hspace{1mm}Daisuke Furihata} \\
	Cybermedia Center\\
	Osaka University\\
 \texttt{daisuke.furihata.cmc@osaka-u.ac.jp}
}

\renewcommand{\headeright}{}
\renewcommand{\undertitle}{}

\begin{document}
\maketitle

\begin{abstract}
	This paper presents a modified neural model for topic detection from a corpus and proposes a new metric to evaluate the detected topics.
The new model builds upon the embedded topic model incorporating some modifications such as document clustering.
Numerical experiments suggest that the new model performs favourably regardless of the document's length.
The new metric, which can be computed more efficiently than widely-used metrics such as topic coherence, provides variable information regarding the understandability of the detected topics. 
\end{abstract}


\section{Introduction}
Topic analysis is a statistical technique used to uncover latent themes within a corpus.
Let $\mathcal{D}$ represent an observed corpus, 
i.e., a collection of documents,
$\mathcal{V}$ a set of observed words,
and $\mathcal{T}$ a collection of known topics.
Each document ${\bm w}_d=(w_{di})_i\in\mathcal{D}$ consists of a sequence of words in $\mathcal{V}$, with $w_{di}$ considered as a realization of a $\mathcal{V}$-valued random variable
$W_{di}$.
Throughout the paper, $d$ and $i$ consistently denote the indices for documents and words, respectively. 
A topic for each word $w_{di}$ is typically modelled by a $\mathcal{T}$-valued random variable $Z_{di}$.
Topic analysis aims to conduct unsupervised inference of a hidden relationship between words and topics modelled as probability distributions, specifically 
$p(W_{di}=v, Z_{di}=t)$.

Latent Dirichlet allocation (LDA)~\cite{bib1} is the most prevalent topic model, which treats 
both the word distribution $p(W_{di}=v \mid Z_{di}=t)$ and topic distribution $p(Z_{di}=t)$ as unknown parameters.   
These parameters are assumed to be drawn from Dirichlet priors and are estimated by using Bayesian inference. 
Although LDA is a powerful model, 
posterior computation is usually expensive.
Furthermore, since LDA relies on word co-occurrence, its performance can be limited in the analysis of short texts due to the lack of frequency information.

In recent years, neural topic models, a new class of models that utilizes deep learning techniques, have been extensively studied. 
These models build upon embeddings and variational inference~\cite{bib7}.
An embedding maps a discrete symbol, such as a word, to a dense vector in a finite-dimensional vector space.  
Specifically, a word embedding refers to an embedding map $\mathcal{E}_{\mathcal{V}}\colon\mathcal{V}\to\mathbb{R}^{H}$, where $H$ is a pre-specified natural number that must be substantially smaller than $\#\mathcal{V}$
(the symbol $\#$ always represents the number of elements of a set). 
Word embedding techniques are incorporated into topic models because they capture the semantic similarity between words more accurately than frequency-based approaches.
Variational inference, an optimization method, is employed in topic models with the expectation of improving the accuracy of the posterior distribution approximation and the model's fit to data.
These techniques are expected to well perform the learning from short texts and can take advantage of parallel computing resources, such as GPUs. 

However, the training process requires considerable trial and error for setting optimization options, such as regularizers, as suitable choices depend heavily on the dataset.
Moreover, unlike classification problems, true labels are not provided in topic analysis, 
making
model evaluation being a long-standing issue~\cite{bib8}.
The most commonly used evaluation metric is topic coherence, among others like perplexity and topic diversity.
Topic coherence measures the semantic consistency within each identified topic.
Nonetheless, the length of each document could influence the metric's value, potentially limiting the reliability of evaluation for relatively short documents.

In light of the two aforementioned issues concerning modelling and evaluation, we shall propose a modified version of neural topic models and also introduce a new metric with the potential to assess the understandability of detected topics.

\section{Preliminaries}
\label{sec2}

Embedded topic model (ETM) treats
both words and topics as unknown embeddings.
ETM models a word distribution as
\begin{equation}
    \label{etmdef}
    p_{\mathcal{E}}(W_{di}=v\mid Z_{di}=t) = \frac{\exp \left[\mathcal{E}_{\mathcal{V}}(v)^{\top}\mathcal{E}_{\mathcal{T}}(t)\right]}{\displaystyle\sum_{v^\prime\in\mathcal{V}}\exp \left[\mathcal{E}_{\mathcal{V}}(v^\prime)^{\top}\mathcal{E}_{\mathcal{T}}(t)\right]},
\end{equation}
where $\mathcal{E}_{\mathcal{T}}:\mathcal{T}\to\mathbb{R}^H$ is a topic embedding.  
The number of topics $\# \mathcal{T}$ is pre-specified, and $H$ is typically set such that $\# \mathcal T < H < \# \mathcal{V}$.
A prior distribution for a topic in a single document is usually set, with the normal distribution, to 
\begin{equation}
    \label{prior_t_normal}
    p(Z_{di}=t\mid {\bm x}_d) = \mathrm{softmax}({\bm x}_d)\mid_{t}, \quad {\bm x}_d \sim \mathrm{N}({\bm 0}, \mathit{I}_{\#\mathcal{T}}).  
\end{equation}
Then, the posterior distribution of the unknown parameter ${\bm x}_d$ reads 
\begin{equation}
    \pi^*_{\mathcal{E}}({\bm x}_1, \dots, {\bm x}_{\#\mathcal{D}}\mid\mathcal{D}) = \frac{1}{\mathcal{M}_{\mathcal{E}}} \left[\prod_{d=1}^{\#\mathcal{D}} p_\mathcal{E}({\bm w}_d\mid{\bm x}_d)\pi_\mathrm{N}({\bm x}_d)\right],
    \nonumber
\end{equation}
where $\pi_\mathrm{N}$ is the probability density function of a standard normal distribution,
\begin{equation}
    \mathcal{M}_{\mathcal{E}} = \int \left[\prod_{d=1}^{\#\mathcal{D}} p_\mathcal{E}({\bm w}_d\mid{\bm x}_d)\pi_\mathrm{N}({\bm x}_d)\right] \mathrm{d}{\bm x}_1\cdots\mathrm{d}{\bm x}_{\#\mathcal{D}}
    \nonumber
\end{equation}
is a marginal likelihood, and
\begin{equation}
    p_{\mathcal{E}}({\bm w}_d\mid {\bm x}_d)
    =
    \prod_{i=1}^{\# {\bm w}_d}
    \sum_{t\in\mathcal{T}}
    p_{\mathcal{E}} (w_{di}\mid t) p(t\mid {\bm x}_d).
    \nonumber
\end{equation}
In the learning process of ETM, we need to infer the embedding $\mathcal{E}_{\mathcal{V}}$ and $\mathcal{E}_{\mathcal{T}}$, and the parameters ${\bm x}_1, \dots, {\bm x}_{\#\mathcal{D}}$ based on the above settings.

Variational inference simplifies the learning process of ETM.
The key concept involves approximating the posterior distribution $\pi_{\mathcal{E}}^\ast$, which is difficult to assess in its current form, by a normal distribution.
The optimization problem is then formulated by using the Kullback--Leibler divergence.

The posterior distribution $\pi_{\mathcal{E}}^\ast$ is approximated by 
\begin{equation}
r_{{\bm \eta}} ({\bm x}_1,\dots,{\bm x}_{\#\mathcal{D}} \mid \mathcal{D}) :=
\prod_{d=1}^{\# \mathcal{D}} r_{{\bm \eta}} ^{(d)}({\bm x}_d\mid {\bm w}_d),
\nonumber
\end{equation}
where each $r_{{\bm \eta}} ^{(d)}$ is the probability distribution function of a normal distribution $\mathrm{N}({\bm m}^{(d)}_{{\bm \eta}}, \mathrm{diag}({\bm s}^{(d)}_{{\bm \eta}})^2)$.
Here, the mean and variance are realized by a neural network $\mathrm{NN}_{{\bm \eta}}^{(1)}\colon \mathcal{D}\to\mathbb{R}^{2\#\mathcal{T}}$ with an unknown parameter ${\bm \eta}$, i.e., 
$\big[({\bm m}^{(d)}_{{\bm \eta}}) ^\top , (\log {\bm s}^{(d)}_{{\bm \eta}}) ^\top \big] ^\top= \mathrm{NN}_{{\bm \eta}}^{(1)}({\bm w}_d)$, 
where $\log{\bm s}_{{\bm \eta}}^{(d)} = (\log s_{{\bm \eta}}^{(d)}(t))_{t\in\mathcal{T}}$.

We then formulate a maximization problem for the following objective function
\begin{equation*}
    \mathcal{L}({\bm \eta},\mathcal{E}_{\mathcal{V}}, \mathcal{E}_{\mathcal{T}}) = -D_{\mathrm{KL}}[r_{{\bm \eta}}\| \pi^*_{\mathcal{E}}] + \log\mathcal{M}_\mathcal{E},
\end{equation*}
where
$D_{\mathrm{KL}}[r_{{\bm \eta}}\| \pi^*_{\mathcal{E}}]$ is a 
Kullback--Leibler divergence.  
This problem is typically solved by gradient descent.  
In solving the problem, the approximation accuracy for the posterior distribution and the model's fit to the data should be improved simultaneously.  
 It should be noted that the word embedding can be kept fixed during training if it has been pre-trained by other models, such as skip-gram. 
 In most topic models, it is reported that pre-training a word embedding improves the performance in short text analysis~\cite{bib5}.  

Topic coherence~\cite{bib10}, the most commonly used evaluation metric in topic analysis, is defined as
\begin{equation}
    \label{tc}
    \mathrm{TC} = \frac{1}{\#\mathcal{T}}\sum_{t\in\mathcal{T}}\left[\frac{1}{N(N-1)}\sum_{\substack{u,v\in U(t,N)\\ u\neq v}} \frac{\log\frac{P(u,v)}{P(u)P(v)}}{-\log P(u,v)}\right],
\end{equation}
where $U(t,N)$ is a set consisting of $N$ words corresponding to the $N$ largest $p_{\mathcal{E}}(v\mid t)$ values, $P(u,v)$ is the probability of words $u$ and $v$ co-occurring in a document, and $P(u)$ is the marginal probability of word $u$.
In general, computing $P(u,v)$ and $P(u)$ for all possible combinations of $u$ and $v$ is highly expensive. 
Following~\cite{bib3}, in the numerical experiments presented below, we will employ the following approximations, which are relatively inexpensive among other approximations, taking in mind that approximation accuracy can be limited:
\begin{align*}
    & P(u,v) \approx \frac{1}{\# \mathcal{D}} 
    \sum_{d=1}^{\# \mathcal{D}}
    \mathbb{I}[u\in \bm{w}_d]\mathbb{I}[v\in \bm{w}_d], \\
    & P(u) \approx \frac{1}{\# \mathcal{D}} 
    \sum_{d=1}^{\# \mathcal{D}}
    \mathbb{I}[u\in \bm{w}_d]
\end{align*}
are distributions of a word in the corpus, where $\mathbb{I}$ is the indicator function.
We note that their definitions depend on $\#\mathcal{D}$.

\section{A modified model}
\label{sec3}

Since neural topic models, including ETM, have a high degree of freedom and admit numerous candidate solutions as local minimum, these models sometimes fail to capture the semantic structure of the corpus. 
To address this issue, training neural topic models necessitates regularization techniques, such as weight decay, dropout, or batch normalization, to guide the model towards better results.
In this section, we introduce three modifications that lead to additional constraints to ETM to facilitate capturing the differences in meaning among documents.
Hereafter, we assume that the corpus $\mathcal{D}$ is partitioned into clusters $C_{1}, \dots, C_{\# \mathcal{T}}$ by using k-means and their centres are denoted by $\boldsymbol{c }(C_{1}), \dots, \boldsymbol{c}(C_{\#\mathcal{T}})$.

First, we replace the topic embedding $\mathcal{E}_{\mathcal{T}}$ with a neural network that inputs cluster centres.
In the case of standard ETM, the output of the topic embedding as $H$-dimensional vectors does not effectively represent the relationship between topics.
Besides, the relationship of the meanings in a document to each other can be measured by computing the inner product similarity (cosine similarity) between cluster centres.
To utilize this prior knowledge, we define a new topic embedding as follows.
Since the number of topics and clusters are the same, $\#\mathcal{T}$, a one-to-one correspondence can be established between topics and clusters.
If a topic $t_j$ corresponds to a cluster $C_j$, we define the topic embedding as
\begin{equation*}
    \tilde{\mathcal{E}}_{\mathcal{T}}(t_j) = \mathrm{NN}^{(2)}_{\boldsymbol{\xi}}(\boldsymbol{c}(C_j)),
\end{equation*}
where $\mathrm{NN}^{(2)}_{\boldsymbol{\xi}}$ is a neural network parametrized by an unknown parameter $\boldsymbol{\xi}$.

Second, we propose a new prior distribution for the parameter $\boldsymbol{x}_d$ based on the clustering outcomes.  
In ETM, the prior distribution for topics is defined by \eqref{prior_t_normal}.
However, this does not incorporate any information about the topic occurrence within the document. 
To represent a prior knowledge that the same topic is likely to appear in documents belonging to the same cluster, we define a new prior distribution as
\begin{equation*}
    \mathrm{N}(\boldsymbol{m}_0^{(d)}, \mathit{I}_{\#\mathcal{T}}), \quad \boldsymbol{m}_0^{(d)} = \lambda^{(d)}\boldsymbol{e}_{n(d)},
\end{equation*}
where $\boldsymbol{\lambda}=\{\lambda^{(d)}\}_{d=1}^{\#\mathcal{D}}$ is a positive unknown parameter, $\boldsymbol{e}_i$ is a unit vector with only the $i$-th component being 1, and $n(d)$ is the cluster number to which the document $\boldsymbol{w}_d$ belongs.

Third, we modify the word probability within a topic based on the word probability in the entire document.
In ETM, since the word distribution in a topic is defined by \eqref{etmdef}, the word-topic similarity $\mathcal{E}_{\mathcal{V}}(v)^{\top}\mathcal{E}_{\mathcal{T}}(t)$ may become large even for words with limited information. 
To circumvent this issue, we modify the word distribution within a topic as follows:
\begin{align*}
    & \tilde{p}_{\mathcal{E},\boldsymbol{\xi}}(W_{di}=v\mid Z_{di}=t)\\
    &\quad = \frac{\exp \left[\mathcal{E}_{\mathcal{V}}(v)^{\top}\tilde{\mathcal{E}}_{\mathcal{T}}(t)\right]G_0(v)}{\displaystyle\sum_{v^\prime\in\mathcal{V}}\exp \left[\mathcal{E}_{\mathcal{V}}(v^\prime)^{\top}\tilde{\mathcal{E}}_{\mathcal{T}}(t)\right]G_0(v^\prime)},
\end{align*}
where 
$G_0 (v) = \sum_{d=1}^{\#\mathcal{D}} \sum_{i=1}^{\#\bm{w}_d} \mathbb{I}[w_{di}=v]\big/ \sum_{d=1}^{\#\mathcal{D}} \# \bm{w}_d$
represents the relative frequency of the word $v\in \mathcal{V}$ in a whole document $\mathcal{D}$.
This modification is expected to make it easier to differentiate between low-information words and words that characterize the topic.

Let $\pi^*_{\mathcal{E}, \boldsymbol{\xi}, \boldsymbol{\lambda}}$ be the posterior distribution of $\boldsymbol{x}_d$ and $\mathcal{M}_{\mathcal{E}, \boldsymbol{\xi}, \boldsymbol{\lambda}}$ be the marginal likelihood in the modified ETM.  Then, the variational inference is a maximization problem with the following objective function:
\begin{align*}
    \mathcal{L}(\boldsymbol{\eta},\mathcal{E}_{\mathcal{V}}, \boldsymbol{\xi}, \boldsymbol{\lambda}) &= -D_{\mathrm{KL}}[r_{\boldsymbol{\eta}}\| \pi^*_{\mathcal{E}, \boldsymbol{\xi}, \boldsymbol{\lambda}}] + \log\mathcal{M}_{\mathcal{E}, \boldsymbol{\xi}, \boldsymbol{\lambda}} \nonumber \\
    & = \sum_{d=1}^{\#\mathcal{D}} \left( \int\left[\log \tilde{p}_{\mathcal{E}, \boldsymbol{\xi}}(\boldsymbol{w}_d\mid \boldsymbol{x}_d)\right]r_{\boldsymbol{\eta}}(\boldsymbol{x}_d\mid\boldsymbol{w}_d)\mathrm{d}\boldsymbol{x}_d- \frac{\#\mathcal{T}}{2} \right. \nonumber \\
    &\phantom{=}\quad -\left.\frac{1}{2}\left(\|\boldsymbol{m}_{\boldsymbol{\eta}}^{(d)} -\boldsymbol{m}_0^{(d)} \|^2 + \|\boldsymbol{s}_{\boldsymbol{\eta}}^{(d)}\|^2 \right) + \sum_{t\in\mathcal{T}}\log s_{\boldsymbol{\eta}}^{(d)}(t) \right).  
\end{align*}

\section{A new metric}
\label{sec4}
In topic modelling, in addition to the coherence, the understandability of the detected topics is also a crucial factor.
We here introduce an evaluation metric that could measure the topic understandability based on the so-called word familiarity\cite{bib9}, which is a measure of the relative understandability of words for humans.  
The higher the word familiarity scores, the more easily understandable the word is. 
Tanaka et al.~\cite{bib9} reported a strong correlation between the word log frequency and word familiarity, with more frequent words being more familiar to people.
In this study, we define a topic understandability metric using the log relative frequency $\log G_0$. 
Intuitively, a topic is deemed easier to understand if the sum or average of $\log G_0$ in the topic is higher. 

For a topic model, we define a weighted sum of word familiarity as
\begin{equation}
    \label{wswf}
    \mathrm{WSWF} = \frac{1}{\mathcal{T}}\sum_{t\in\mathcal{T}}\left[\sum_{v\in U(t, N)}\hat{p}(v\mid t)\log G_0 (v)\right],
\end{equation}
where $\hat{p}(v\mid t)$ is the estimated word distribution in the topic $t$.  
We consider the model with a higher value of WSWF to have more understandable topics.
We note that WSWF can be computed instantly once $\hat{p}$ is established.

\section{Numerical experiment}
\label{sec5}
We study the following corpora.    
\begin{enumerate}
\item 20News--Groups:\\ {\sf http://qwone.com/\url{~}jason/20Newsgroups\\ /20news-18828.tar.gz} ~(accessed on 9 January 2023)
\item Search--Snippets:\\ {\sf http://jwebpro.sourceforge.net/data-web-\\ snippets.tar.gz} ~(accessed on 9 January 2023)
\end{enumerate}
Table~\ref{table:data} lists the details of each dataset. 
20--NewsGroups is a standard dataset in NLP community and consists of multiple news articles in each of 20 categories. On the other hand, Search--Snippets is a collection of related keywords for each Wikipedia article and the length of each document is relatively short. 
As preprocessing, numbers, special characters, pronouns, prepositions, and low-frequent words were removed. 
In addition, words were lowercased and lemmatized.

\begin{table}[t]
    \vspace*{-6pt}
    \caption{Information for each dataset.  \label{table:data} }
    \vspace*{3pt}
  \centering
  \begin{tabular}{cccc}
    \hline
    &$\#\mathcal{D}$&$\#\mathcal{V}$&Avg. length\\
    \hline
    20--NewsGroups&18827&7558&81.9\\
    Search--Snippets&10059&9577&15.8\\
    \hline
  \end{tabular}
\end{table} 

We compared five models with $\#\mathcal{T}=50$: LDA, ETM, ETM with a pre-trained word embedding, our modified model and our modified model with a pre-trained word embedding.  We pre-trained the word embedding using the skip-gram model~\cite{bib6} for each corpus with the embedding dimension $H=200$. 
We evaluated all the models by topic coherence and WSWF with $N=10$. 
Additionally, we provided subjective accuracy for the 20-NewsGroups dataset. 
We compared the content of $100$ randomly sampled documents 
with the top $10$ most probable words from the predictive distribution for each document.
The evaluation was conducted in a subjective manner based on the consistency of the words, the presence of keywords in the document, and the effectiveness of the summary.
Since each document in the Search-Snippets dataset consists of a list of keywords, making the comparative evaluation difficult, subjective evaluation was not performed.

\begin{table}[t]
    \vspace*{-6pt}
    \caption{Model evaluation in 20--NewsGroups.Higher is better.  The row pretrained indicates whether the pretrained word embedding is Available or NOT.  Subj. indicates subjective accuracy.  \label{table:tngevaltable} }
    \vspace*{3pt}
  \centering
  \begin{tabular}{p{2.5cm}p{2.5cm}p{2.5cm}p{2.5cm}p{2.5cm}p{2.5cm}}
    \hline\hline
     \multicolumn{1}{c}{model} & \multicolumn{1}{c}{LDA} & \multicolumn{2}{c}{ETM} & \multicolumn{2}{c}{modified}\\
     
     \multicolumn{1}{c}{pretrained} & \multicolumn{1}{c}{} & \multicolumn{1}{c}{N} & \multicolumn{1}{c}{A} & \multicolumn{1}{c}{N} & \multicolumn{1}{c}{A}\\
    \hline
    \multicolumn{1}{c}{TC} & \multicolumn{1}{c}{\bf 0.274} & \multicolumn{1}{c}{0.085} & \multicolumn{1}{c}{0.232} & \multicolumn{1}{c}{0.245} & \multicolumn{1}{c}{0.254}\\
    \multicolumn{1}{c}{WSWF} & \multicolumn{1}{c}{\bf -0.833} & \multicolumn{1}{c}{-2.867} & \multicolumn{1}{c}{-1.793} & \multicolumn{1}{c}{-1.053} & \multicolumn{1}{c}{-0.962} \\
    \multicolumn{1}{c}{Subj.~[\%]} & \multicolumn{1}{c}{\bf 55} & \multicolumn{1}{c}{4} & \multicolumn{1}{c}{44} & \multicolumn{1}{c}{54} & \multicolumn{1}{c}{\bf 55} \\
    \hline\hline
  \end{tabular}
\end{table} 

\begin{figure}[t]
    \centering \includegraphics[scale=0.48]{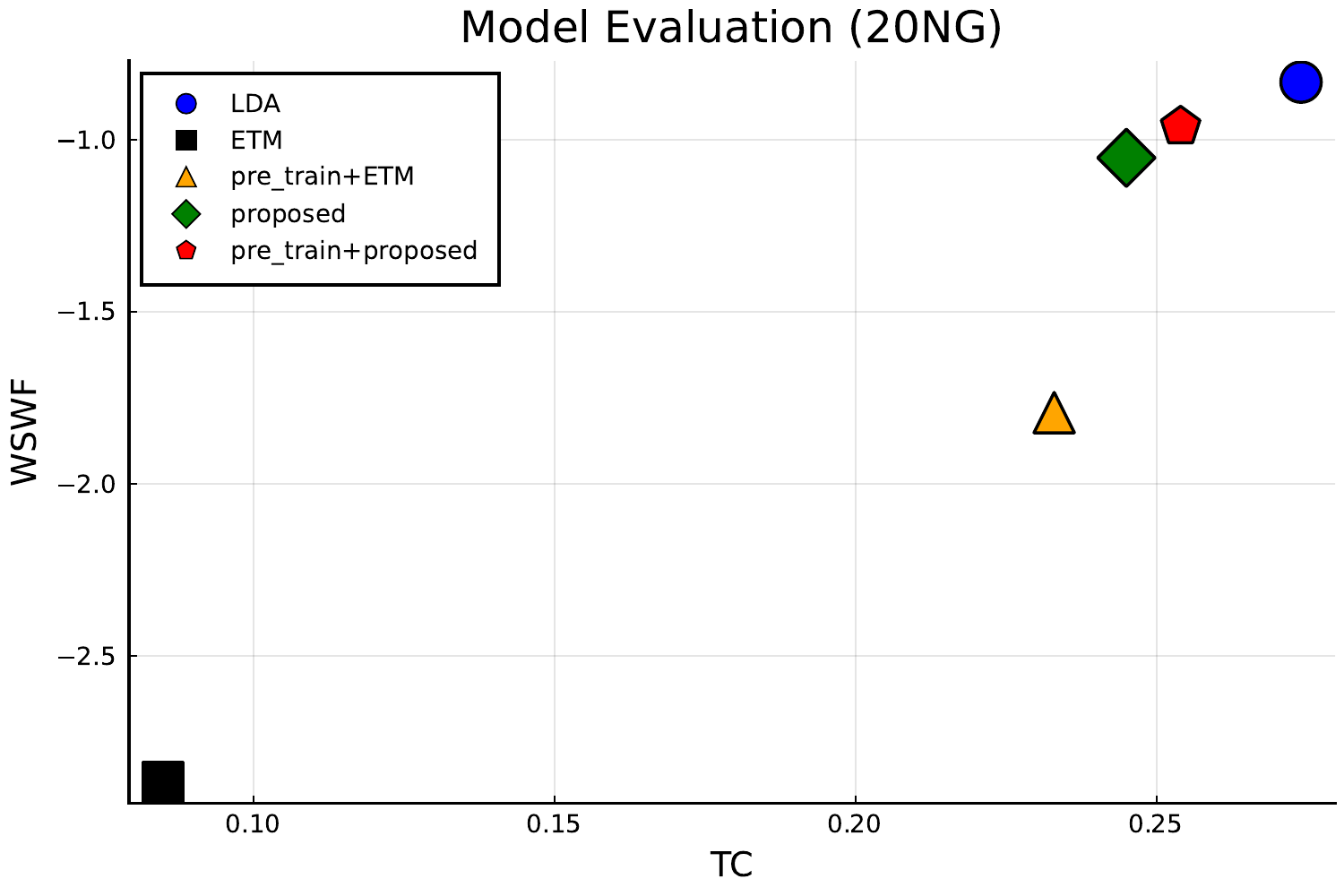} 
    \vspace*{-9pt}
    \caption{Model comparison by topic coherence and WSWF in 20--NewsGroups. Higher is better.\label{TNG}}
\end{figure}

Table~\ref{table:tngevaltable} lists the value of each evaluation metric for 20-NewsGroups, and Fig.~\ref{TNG} visualize the comparison between TC and WSWF for each model.
We observe that all models except for ETM without pre-trained word embedding achieve high scores. 
In addition, whether or not the pre-trained word embedding is incorporated, our modified model outperforms ETM in both metrics, which indicates that the modification improves the topic coherence and understandability simultaneously for this dataset.
Furthermore, our modified model achieves almost the same level as LDA in subjective evaluation.

\begin{figure}[t]
    \centering \includegraphics[scale=0.48]{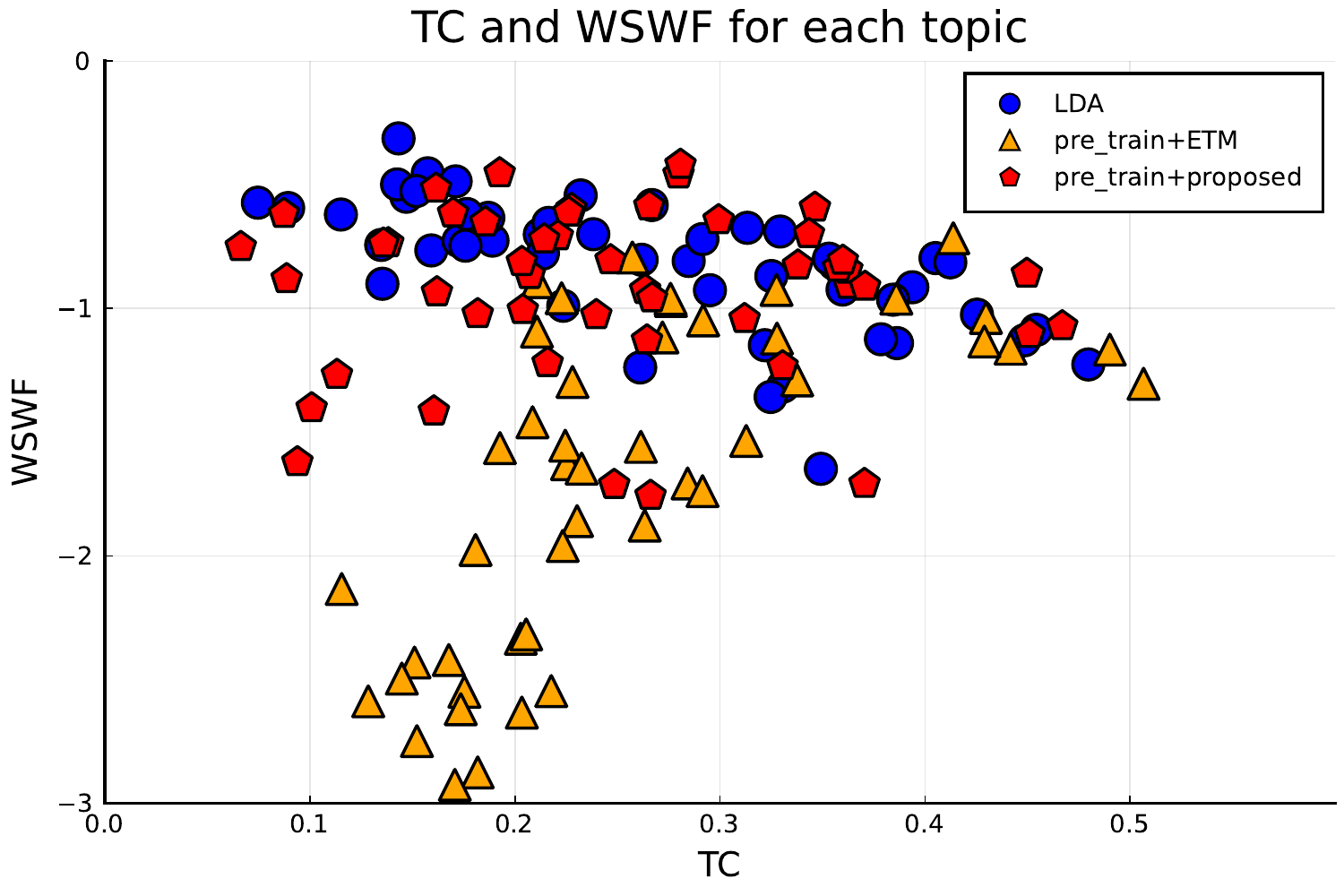} 
    \vspace*{-9pt}
    \caption{Plot of TC and WSWF pairs for each topic detected by LDA (blue circles) and ETM with word embedding pretraining (orange triangles), respectively. \label{LDAvspETM}}
\end{figure}

Note that both TC \eqref{tc} and the proposed metric WSWF \eqref{wswf} are defined as the average of the ratings for each topic. 
However, comparing the values of TC and WSWF before averaging may give us some insights into the quality of the detected topics.
To illustrate this, we depict the values of LDA, ETM with pre-training
and the modified model with pre-training in Fig.~\ref{LDAvspETM} (the modified model without pre-training exhibited similar results that with pre-training).
This figure suggests that topics detected by LDA tend to consist of more common words, while those by ETM are more likely to comprise less familiar terms, such as technical jargon.
In contrast, the modified model detects topics with a higher proportion of familiar words than ETM.

\begin{table}[t]
    \vspace*{-6pt}
    \caption{Model evaluation in Search--Snippets.Higher is better.  The row pretrained indicates whether the pretrained word embedding is Available or NOT.\label{table:wssevaltable} }
    \vspace*{3pt}
  \centering
  \begin{tabular}{p{2.5cm}p{2.5cm}p{2.5cm}p{2.5cm}p{2.5cm}p{2.5cm}}
    \hline\hline
     \multicolumn{1}{c}{model} & \multicolumn{1}{c}{LDA} & \multicolumn{2}{c}{ETM} & \multicolumn{2}{c}{modified}\\
     
     \multicolumn{1}{c}{pretrained} & \multicolumn{1}{c}{} & \multicolumn{1}{c}{N} & \multicolumn{1}{c}{A} & \multicolumn{1}{c}{N} & \multicolumn{1}{c}{A}\\
    \hline
    \multicolumn{1}{c}{TC} & \multicolumn{1}{c}{-0.191} & \multicolumn{1}{c}{-0.659} & \multicolumn{1}{c}{-0.336} & \multicolumn{1}{c}{-0.115} & \multicolumn{1}{c}{\bf 0.014}\\
    \multicolumn{1}{c}{WSWF} & \multicolumn{1}{c}{\bf -1.590} & \multicolumn{1}{c}{-3.927} & \multicolumn{1}{c}{-2.528} & \multicolumn{1}{c}{-1.830} & \multicolumn{1}{c}{-1.723} \\
    \hline\hline
  \end{tabular}
\end{table}

Table~\ref{table:wssevaltable} displays the values of each metric for Search--Snippets.
LDA, a frequency-based model, shows law coherence because of the lack of co-occurrence information. In contrast, the modified model exhibits the highest coherence. 
This observation suggests that the modified model seems effective even when the length of each document is relatively short.

\section{Discussions}

In this paper, we have introduced the modified model for topic detection and proposed a new metric, WSWF, that measures the understandability of the detected topics.
Numerical experiments suggest that the new model, particularly with pre-training, outperforms ETM irrespective of the length of each document.
Further, WSWF is efficiently computed, independently of the length of each document.
If the user prioritizes understandability, the new metric could be highly beneficial. 
We also emphasize that examining both TC and WSWF could offer variable insights into the quality of detected topics.

\end{document}